\documentclass[letterpaper, 10 pt, conference]{ieeeconf}  

\IEEEoverridecommandlockouts                              

\usepackage{graphicx}
\usepackage{booktabs}
\usepackage{amsmath} 

\usepackage{array}     
\usepackage{longtable} 
\usepackage{tabularx}  
\usepackage{ragged2e}  
\usepackage{tabularray}
\usepackage{float} 
\usepackage{color}
\usepackage{multirow}
\usepackage{dblfloatfix}   
\usepackage{placeins} 
\usepackage{capt-of}
\usepackage{cuted} 
\usepackage{balance} 

\usepackage[bookmarks=true]{hyperref} 
    \hypersetup{colorlinks=true, linkcolor=black, urlcolor=blue, citecolor=black} 
\usepackage{cleveref} 
\usepackage{multirow}
\usepackage{booktabs}
\Crefformat{figure}{#2Fig.~#1#3} 

\usepackage[colorinlistoftodos]{todonotes}
\let\oldtodo\todo
\renewcommand{\todo}[1]{\oldtodo[inline, color=red!25]{#1}}

\title{\LARGE \bf
A Benchmark of Dexterity for Anthropomorphic Robotic Hands
}

\author{Davide Liconti$^{1 *}$, Yuning Zhou$^{1 *}$, Yasunori Toshimitsu$^{1}$, Ronan Hinchet$^{1}$ and Robert K. Katzschmann$^{1}$
\thanks{*These authors contributed equally to this work.}
\thanks{$^{1}$Soft Robotics Lab, D-MAVT, ETH Zurich
        {\tt\small \href{mailto:rkk@ethz.ch}{rkk@ethz.ch}}}%
}

\begin{document}

\maketitle
\thispagestyle{empty}
\pagestyle{empty}

\vspace{-1.5em}


\begin{strip}
\vspace{-3.5em}
\centering
\includegraphics[width=\textwidth]{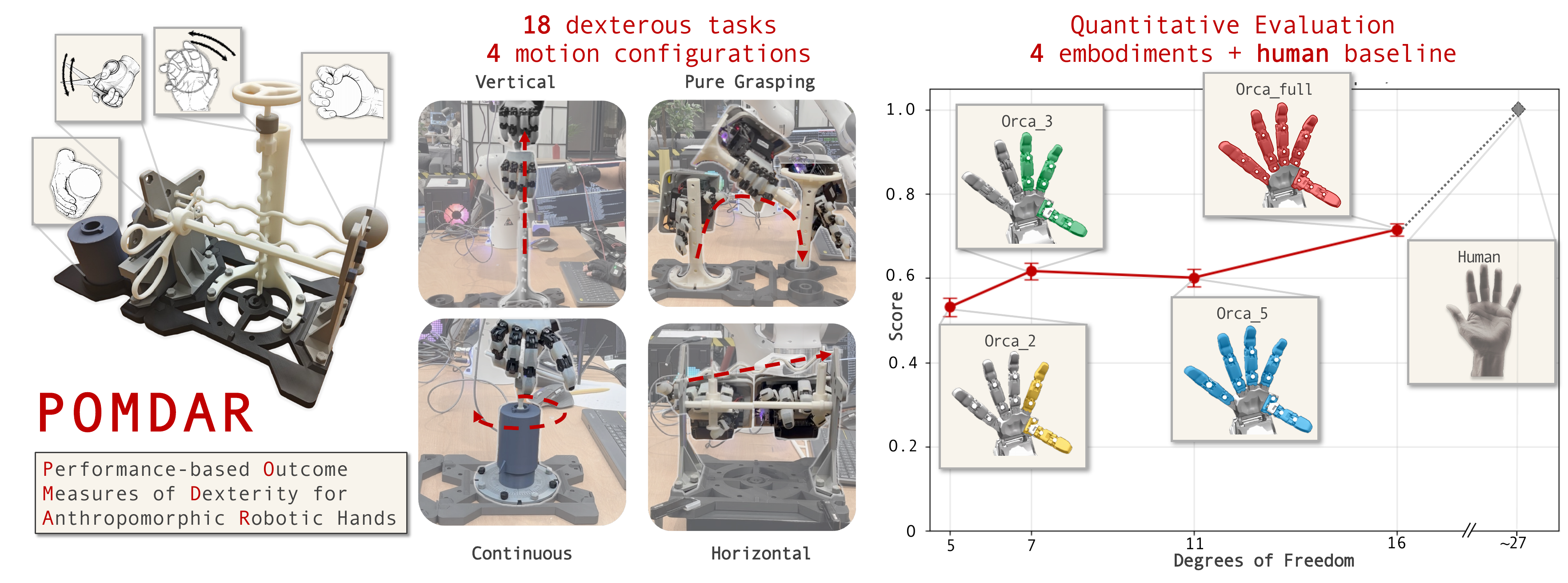}
\captionof{figure}{POMDAR (Performance-based Outcome Measures of Dexterity for Anthropomorphic Robot Hands): a compact, fully 3D-printable benchmark for quantitative evaluation of robotic hand dexterity. The setup includes four manipulation configurations: vertical and horizontal scaffolded configurations, continuous rotation, and pure grasping, covering a wide range of dexterous skills. Quantitative results obtained via teleoperation with ORCA hands of increasing DoF (2 to 16) show improved performance with higher embodiment complexity, demonstrating the benchmark’s sensitivity to dexterity variations.}
\label{fig:teaser}
\vspace{-0.8em}
\end{strip}

\begin{abstract}
Dexterity is a central yet ambiguously defined concept in the design and evaluation of anthropomorphic robotic hands. In practice, the term is often used inconsistently, with different systems evaluated under disparate criteria, making meaningful comparisons across designs difficult. This highlights the need for a unified, performance-based definition of dexterity grounded in measurable outcomes rather than proxy metrics.
In this work, we introduce POMDAR, a comprehensive dexterity benchmark that formalizes dexterity as task performance across a structured set of manipulation and grasping motions. The benchmark was systematically derived from established taxonomies in human motor control. It is implemented in both real-world and simulation and includes four manipulation configurations: vertical and horizontal configurations,
continuous rotation, and pure grasping. The task designs contain mechanical scaffolding to constrain task motion, suppress compensatory strategies, and enable metrics to be measured unambiguously.
We define a quantitative scoring metric combining task correctness and execution speed, effectively measuring dexterity as throughput. This enables objective, reproducible, and interpretable evaluation across different hand designs. POMDAR provides an open-source, standardized, and taxonomy-grounded benchmark for consistent comparison and evaluation of anthropomorphic robot hands to facilitate a systematic advancement of dexterous manipulation platforms.
CAD, simulation files, and evaluation videos are publicly available at \url{https://srl-ethz.github.io/POMDAR/}.
\end{abstract}


\section{Introduction}
\label{sec:Introduction}


For anthropomorphic robotic hands, \textit{dexterity} is not only a central design objective but also the defining attribute that distinguishes them from traditional grippers. Yet a major research gap persists: there is no unified definition of dexterity for anthropomorphic hands, nor a standardized framework for evaluating it across designs. This limits meaningful comparison across systems and hinders the community’s ability to systematically design, optimize, and select hands for specific manipulation tasks.

A key challenge is that dexterity cannot be captured through kinematic properties alone. Measures such as degrees of freedom, joint limits, or manipulability indices reflect a system’s \emph{potential} but not how effectively a hand performs real manipulation under contact-rich interactions. Dexterity must therefore be evaluated in a \textit{performance-based} manner, which requires specifying a representative set of tasks grounded in established studies of hand motion, grasping, and manipulation primitives.

The consequences of this dexterity definition and evaluation gap are visible even in top venues. Two anthropomorphic hands recently published in \textit{Nature Communications}, the \textit{SMA Hand}~\cite{yang_lightweight_2025} and the \textit{ILDA Hand}~\cite{kim_integrated_2021}, share only one common metric: finger reachable workspace. All other assessments rely on ad-hoc measures such as specific grasp postures or custom-designed tasks.

More broadly, the lack of a standardized benchmarking framework~\cite{falco_grasping_2015} means that researchers, developers, and end users have no reliable method to compare dexterity across hand designs~\cite{coulson_elliott_2021, zhou_50_2020, falco_grasping_2015}. This complicates the assessment of design improvements and limits the ability to match hand designs to specific task requirements.

To address these gaps, we introduce POMDAR, a systematic dexterity benchmarking framework applicable to both physical and simulated environments. Our benchmark is guided by the following design principles:
\begin{enumerate}
\item \textbf{Representative of real hand use}: Every benchmark task can be traced to an existing manipulation or grasp taxonomy, ensuring a representative choice of tasks.
\item \textbf{Reproducible across laboratories}: The benchmark design is open source and can be fully 3D printed, ensuring easy access without external procurement. Many components are reused across tasks, further facilitating fabrication. In addition, the benchmark is also available in simulation, allowing for pre-fabrication dexterity evaluation and design optimization.
\item \textbf{Directly observable, standardized motions}: The benchmark uses mechanical scaffolds to constrain task motion to the intended degrees of freedom to suppress compensatory strategies (e.g., gravity assistance, palm support, or excessive arm/wrist involvement), standardizing movements and enabling direct observation of the task completion level.
\item \textbf{Quantitative, throughput-based evaluation}: Dexterity is measured through a unified score that combines task success and execution speed, effectively capturing performance as task throughput. This enables objective, continuous, and easily comparable evaluation across different hand designs.
\end{enumerate}
\section{Approaches to define and measure dexterity}
\label{sec:background}

\begin{figure*}
        \centering
        \includegraphics[width=1.0\linewidth]{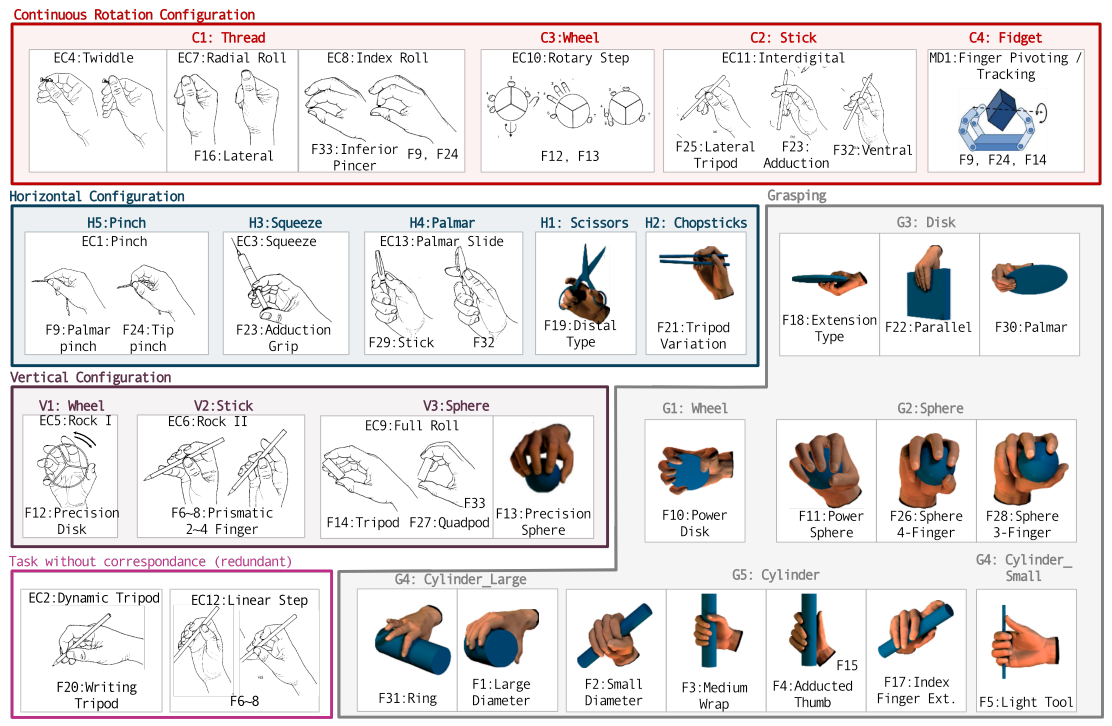}
        \caption{Overview of hand tasks based on those identified in Elliott \& Connolly's taxonomy for manipulation \cite{elliott_classification_1984}, Ma \& Dollar's extension~\cite{ma_dexterity_2011}, and Feix's GRASP taxonomy \cite{feix_grasp_2016}, respectively identified with the symbols \texttt{EC}, \texttt{MD}, and \texttt{F}. Many of the GRASP taxonomy postures are inherently contained in each of the manipulation patterns, providing an opportunity to make the benchmark more efficient. The taxonomies are grouped in the POMDAR benchmark task. Refer to \Cref{fig:benchmark_overview} for the actual implementation of the tasks.}
        \label{fig:manipulation_grasp_classification}
\end{figure*}
\subsection{Definition of Dexterity}
\label{subsec:def_dexterity}

Early medical literature distinguished between \emph{fine} (finger) dexterity, involving coordinated finger movements on smaller objects, and \emph{gross} (manual) dexterity, involving whole-hand and arm movements on larger objects~\cite{fleishman_factor_1954}. Later rehabilitation definitions increasingly characterized dexterity as coordinated voluntary movement for functional object manipulation, emphasizing speed and task completion~\cite{backman_assessment_1992}.

More recent formulations extend the concept to include adaptation to environmental changes and task demands, framing dexterity as an interaction between motor control and external constraints~\cite{elangovan_analysis_2022}. This implies that dexterity is inherently tied to dynamic behavior and cannot be measured from static grasps alone.

In robotics, Cutkosky~\cite{cutkosky_grasp_1989} argued that dexterity depends on two parameters: manipulability (the ability of a grasp to impart arbitrary motions to an object) and kinematic workspace. Bicchi~\cite{bicchi_hands_2000} later emphasized in-hand object repositioning and reorientation which are enabled by kinematic redundancy.

\subsection{Classification of Dexterous Tasks}
Researchers in both medicine and robotics have proposed taxonomies of human hand activities. Bullock et al.~\cite{bullock_hand-centric_2013} provide a compact, motion-centric map of manipulation using five binary descriptors (contact, prehensile/non-prehensile, motion, within-hand motion, motion-at-contact), yielding 15 mutually exclusive classes. From these, grasping (static contact to constrain an object) and in-hand manipulation (dynamic contact for repositioning or reorienting an object) emerge as two principal categories.

For \emph{manipulation}, Elliott and Connolly~\cite{elliott_classification_1984} contribute a human-derived set of 13 coordination patterns (synergy- and sequence-based), forming a minimal library of \emph{dexterity primitives} for task selection. Ma and Dollar~\cite{ma_dexterity_2011} augment this set with robotics-relevant categories (regrasping, in-grasp manipulation, finger gaiting, rolling, sliding, and finger pivoting/tracking), explicitly covering pivoting/tracking behaviors that are otherwise underrepresented.

For \emph{grasping}, Napier's power--precision split~\cite{napier_prehensile_1956} distinguishes force-dominant from dexterity-dominant grasps. Cutkosky~\cite{cutkosky_grasp_1989} adds a task-driven hierarchy based on object geometry, clamping, and force. Feix et al.'s GRASP Taxonomy~\cite{feix_grasp_2016} provides a comprehensive set of 33 stable, one-handed prehensile grasp types, classified by opposition type, virtual-finger roles, and thumb position, forming a practical catalog for selecting standardized grasping postures.

\section{Existing Dexterity Benchmarks}
\label{sec:benchmarks_and_pomdar}
In the literature, existing evaluation practices broadly fall into two families: \emph{parameter-based} (proxy) evaluation and \emph{task-performance-based} evaluation, the latter aligning with \textit{performance-based outcome measures of dexterity} (PBOMD) used in clinical assessment \cite{yong_performance-based_2022}.

\subsection{Parameter-based (proxy) evaluation paradigm}
\label{subsubsec:parametric_eval_paper}
A common practice in both academic reporting and commercial specifications is to infer ``dexterity'' from design parameters such as the number of degrees of freedom, joint ranges of motion, actuator count, speed, and kinematic indices (e.g., Jacobian-based manipulability and conditioning) \cite{elangovan_analysis_2022}.
Such measures implicitly target \emph{potential} capability rather than \emph{realized} capability in contact-rich manipulation.
Although such evaluation is fast to compute and inexpensive, it does not correspond to a concrete task set and does not directly evaluate grasp stability, contact transitions, or control in actual physical interaction.

\subsection{Task-performance-based evaluation (PBOMD-style) paradigm}
\label{subsubsec:PBOMD_eval_paper}
PBOMDs, widely adopted in clinical settings~\cite{yong_performance-based_2022}, evaluate dexterity by measuring how effectively a subject (human or robot) performs physical tasks. Rather than relying on abstract mathematical constructs, PBOMDs use quantitative performance metrics such as task completion time, error rates, or motion efficiency to provide practical, outcome-driven assessments of manual dexterity.

In the robotics context, task-performance-based benchmarks assess dexterity by measuring whether and how well a hand accomplishes a set of physical tasks (e.g., stable grasping, in-hand object reorientation, sequential finger gaiting), using outcome metrics such as success rate, achieved displacement/rotation, or completion time \cite{coulson_elliott_2021, zhou_50_2020, elangovan_accessible_2022}. Compared to parameter-based proxies, PBOMD-style benchmarking improves face validity and typically yields more actionable comparisons for end users.

\subsubsection{The Elliott \& Connolly Benchmark (E\&C)}
\label{subsubsec:ec_benchmark_paper}
Coulson \emph{et al.}~\cite{coulson_elliott_2021} proposed using the E\&C taxonomy~\cite{elliott_classification_1984} to specifically isolate \emph{in-hand dexterity} of humanoid robot hands.

E\&C focuses on \emph{within-hand} object motions (translations and rotations relative to the hand) achieved through digit motion alone, without the arm, wrist, or external support~\cite{coulson_elliott_2021}.
The benchmark targets core in-hand primitives such as rolling, sliding, and sequential repositioning (finger-gaiting-like patterns).
Object selection largely leverages the YCB set, supporting standardization, although its procurement has become increasingly difficult.
However, it does not include pure grasping tasks (static grasp formation and stability), and it typically relies on visual tracking (e.g., AprilTags), which adds instrumentation burden and can suffer from occlusions during manipulation.

\subsubsection{50 Hand Dexterity Benchmarks (HD-marks)}
\label{subsubsec:hdmarks_paper}
Zhou \emph{et al.} introduced a broad checklist-style benchmark containing 50 tasks spanning grasping, thumb dexterity, and in-hand manipulation \cite{zhou_50_2020}.
The benchmark includes: (i) a large set of static grasps drawn from the GRASP taxonomy \cite{feix_grasp_2016}, (ii) thumb postures inspired by the Kapandji test \cite{kapandji_cotation_1986}, and (iii) a small set of in-hand tasks parameterized by translations/rotations along Cartesian axes \cite{zhou_50_2020}.
Its primary strength is breadth: it covers grasp diversity, thumb mobility, and basic in-hand motion directions within a single framework~\cite{zhou_50_2020}.
However, evaluation is largely binary (success/failure), and the benchmark does not enforce standardized objects across tasks, reducing cross-lab comparability.

\subsubsection{Dexterity Test Board}
\label{subsubsec:dexterity_board_paper}
Elangovan \emph{et al.} proposed a modular dexterity test board comprising 24 tasks across multiple categories (simple manipulation, reorientation, fine manipulation, fastening tasks, and puzzles) \cite{elangovan_accessible_2022}.
The motorized board includes pick-and-place tasks, object reorientation, threaded insertions/turning, nut/bolt tool interactions, and puzzle-like force/control tasks, with an open, modular construction that promotes replication~\cite{elangovan_accessible_2022}.
Although the tasks were adapted from existing dexterity tests, they are not mapped one-to-one to prior tests or to a formal taxonomy, making the theoretical foundation weaker than that of the other benchmarks.

\section{Construction of benchmark tasks for POMDAR}
To cover the full breadth of hand capabilities, we draw on 14 manipulation patterns for in-hand manipulation and 33 grasp types from the GRASP Taxonomy~\cite{feix_grasp_2016} as the basis for constructing benchmark tasks. The 14 manipulation patterns are derived from Elliott and Connolly’s taxonomy of 13 patterns~\cite{elliott_classification_1984} (where ``Rock’’ is split into two variants~\cite{coulson_elliott_2021}), plus Finger Pivoting/Tracking from Ma and Dollar~\cite{ma_dexterity_2011}. We translate these taxonomies into a set of self-contained, concrete tasks that constitute the benchmark.

As shown in the upper portion of \Cref{fig:manipulation_grasp_classification}, many grasp patterns are inherently contained within the manipulation patterns: 16 of the 33 grasp types already appear. This provides an opportunity to streamline the benchmark by evaluating manipulation and grasping simultaneously. The remaining grasp patterns, shown in the lower portion of \Cref{fig:manipulation_grasp_classification}, do not overlap with any manipulation motion.

Based on this observation, a dual-structured task categorization is proposed for the comprehensive dexterity benchmark:
\begin{itemize}
\item \textbf{Manipulation Tasks}: Motion-based tasks that primarily measure in-hand dexterity based on Elliott and Connolly's manipulation taxonomy. Because these tasks involve both object interaction and configuration transitions, they are well suited for simultaneously evaluating grasping and in-hand manipulation performance.
\item \textbf{Pure Grasping Tasks}: These tasks specifically evaluate the grasp types that do not overlap with in-hand manipulation patterns (shown in the lower part of \Cref{fig:manipulation_grasp_classification}). They primarily assess the hand's ability to form and maintain stable grasp configurations.
\end{itemize}

This task division aligns closely with the observation of dexterity seen in medical literature \cite{fleishman_factor_1954}, which distinguishes between:
\begin{itemize}
    \item \textit{Gross Dexterity}, defined as the coordinated action between fingers and palm, and primarily reflected in \textbf{pure grasping tasks};
    \item \textit{Fine Dexterity}, defined as inter-digit coordination during object interaction, and best assessed through \textbf{ manipulation tasks}.
\end{itemize}
Such a structured task design ensures both comprehensiveness and efficiency, and reflects the multidimensional nature of dexterity in anthropomorphic robotic hands.

\begin{figure*}
        \centering
        \includegraphics[width=1.0\linewidth]{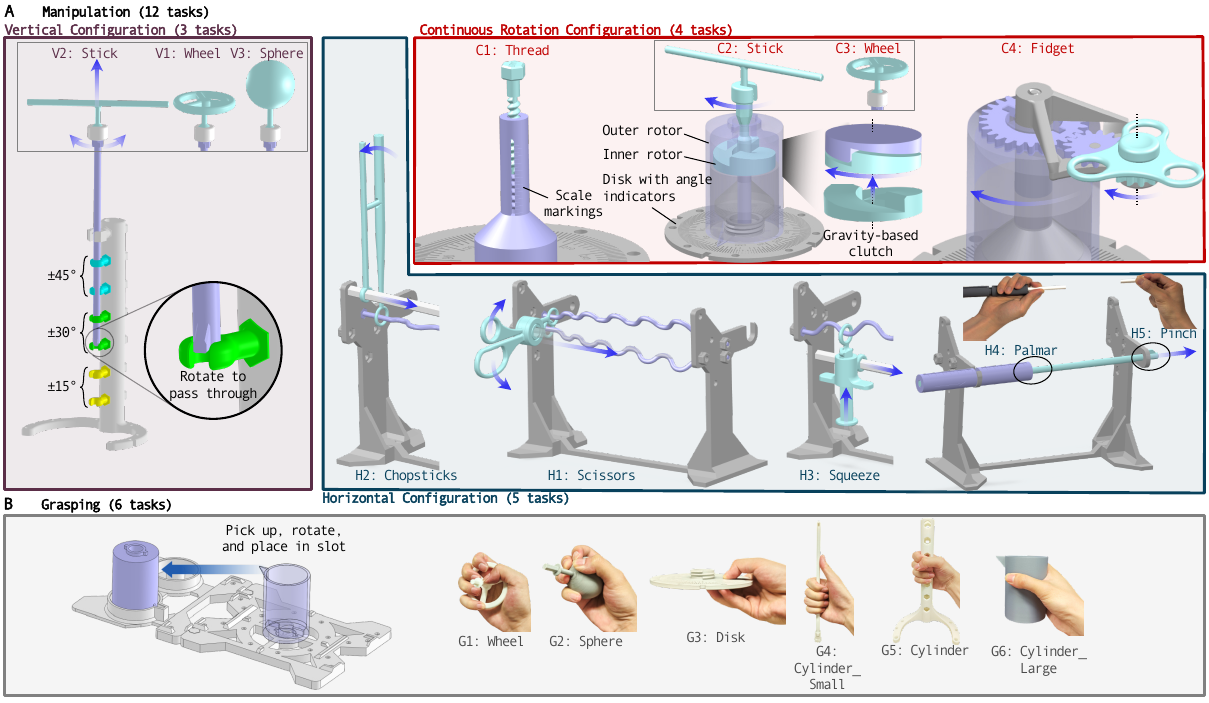}
    \caption{The benchmark comprises four task configurations: two scaffolded manipulation setups (vertical and horizontal), a continuous rotation configuration, and a set of pure grasping tasks. In the vertical scaffolded configuration (V), the hand grasps an object and moves it upward along a rod with discrete notches of increasing angle (from $\pm 15^\circ$ to $\pm 45^\circ$), requiring coordinated in-hand adjustments to pass each constraint. 
    In the horizontal scaffolded configuration (H), the object is translated along a curved rail with progressively increasing curvature, enforcing controlled in-hand manipulation to clear each section. The number of curves depends on the task: three for the chopsticks task (H5), four for the squeeze task (H2), and six for the scissors task (H1). The palmar (H3) and pinch (H4) tasks share the same object but involve manipulating different parts of it. The setup uses a horizontal hexagonal rod to constrain motion along the horizontal axis. For the chopsticks task, the two sticks undergo a constrained motion of approximately $\pm 20^\circ$. 
    The continuous rotation configuration (C) features a gravity-based clutch mechanism that engages the outer rotor to maintain object suspension, enabling continuous, stepwise rotation without resetting (e.g., C2: stick, C3: wheel). The fidget task uses a simple geared transmission to transfer motion to a rotating element, while the thread task consists of a plastic screw that must be removed from a threaded hole. 
    The pure grasping tasks (G) isolate grasp quality by requiring the hand to pick up objects from inserts and relocate them in free space, without external scaffolding. This avoids artificially stabilizing suboptimal grasps, enforcing more robust grasping. A total of six grasp objects are used. Many objects are reused across configurations to ensure consistency and reduce redundancy in the benchmark design. The setup is compact (see Fig.~\ref{fig:teaser}), fully 3D printable, requires no additional components beyond printing material, and can be fixed to a table using clamps.}        \label{fig:benchmark_overview}
\end{figure*}

\subsection{Consolidating manipulation tasks}
\label{sec: manipulation tasks consolidation}
The proposed benchmarking system is designed to be performance-based and outcome-oriented, and should avoid direct penalization of robotic hands with unconventional structures, such as those with fewer fingers.
Thus, the prismatic 2--4 finger patterns used for precision grasping (F6 to F8) were consolidated into a single representative manipulation task \textit{V2: Stick} (\Cref{fig:benchmark_overview}), which rotates a thin object in its fingers.

The grasping patterns not assigned to manipulation tasks in \Cref{fig:manipulation_grasp_classification} are mostly focused on firmly securing objects in the hand, with two notable exceptions: F19 (Distal Type) and F21 (Tripod Variation), typically demonstrated through the use of scissors and chopsticks, respectively. Because these patterns entail functional within-hand motions that resemble in-hand manipulation rather than static grasping, they were incorporated into the \textbf{Manipulation Tasks} category.

Tripod Variation (F21) is highly similar to Dynamic Tripod (EC2, F20), which describes writing with a pen. In both patterns, the index finger rests on the top surface of the pen or upper chopstick, while the middle finger and thumb press against the side surfaces, forming a stable three-point contact. The key distinction is that F21 requires simultaneously holding an additional chopstick. Given this similarity, F21 was integrated into \textit{H2: Chopsticks}, while G19 (Distal Type) was renamed \textit{H1: Scissors}.
EC12 (linear step) is a variation of EC13, but using the grap of EC6, so it was considered redundant and not implemented in the benchmark.

The tasks EC4, EC7, and EC8 all require using a combination of thumb and index flexion to rotate a very small object. Since it is difficult to implement tasks that test all these different manipulation strategies, and we aim to obtain quantitative and objective performance measures (rather than subjective evaluation of how closely the intended motion is followed), all these tasks were incorporated into the \textit{C1: Thread} task.
The motion of EC9 (full roll), using the F13 precision sphere grasp, was grouped into a vertical configuration task \textit{V3: Sphere}, with the goal of reusing as many components as possible while maintaining a compact setup.

This results in a total of 12 manipulation tasks (\Cref{fig:benchmark_overview}).

\subsection{Consolidating grasping tasks}
\label{sec: grasping tasks consolidation}

To make the benchmark fairer to robot hands with fewer fingers and as objective as possible, a similar rationale applies to the grasping patterns. For the grasping tasks, since we aim to avoid non-objective scoring and instead rely purely on performance-based evaluation, we cluster the grasps based on the manipulated object. Specifically, we group F2, F3, F4, F15, and F17 into cylindrical grasps, with variations including F1 and F31 for large-diameter cylinders, and F5 for small-diameter cylinders.

We then consider grasping of spherical objects (F26, F28) and disk-shaped objects (F18, F22, F30). In total, six grasp tasks are evaluated. An overview of the final grasp tasks is shown in \Cref{fig:benchmark_overview}B.


\subsection{Design of the POMDAR benchmark}
We designed the POMDAR benchmark tasks to fulfill the design principles introduced in \Cref{sec:Introduction}, as illustrated in \Cref{fig:benchmark_overview}. 
The manipulation tasks (\Cref{fig:benchmark_overview}A) are further organized into three configurations:
\begin{itemize}
\item \textbf{Vertical (scaffolded)} configuration, where the hand grasps an object and wiggles it while pulling a rod upward through a sequence of notches.
\item \textbf{Continuous Rotation} configuration, where the hand grasps and continuously rotates an object through stepping sequences. A gravity-based clutch engages the outer rotor to keep the object suspended. This platform can also be adapted for tasks such as fidget spinning and threading.
\item \textbf{Horizontal (scaffolded)} configuration, where the object is manipulated in-hand while being translated horizontally. The required range of motion increases progressively to clear a curved rail, thereby increasing task difficulty. This setup can also be used for manipulating thin rods.
\end{itemize}
The pure grasping tasks (\Cref{fig:benchmark_overview}B) reuse components from the manipulation setups and measure grasps not inherently covered by the manipulation tasks. In these tasks, objects are not constrained by external structures and must be stably held while undergoing rotations in free space, enabling evaluation of grasp robustness and quality.
\subsection{Benchmark protocol and scoring}

The POMDAR benchmark quantifies dexterity as the \textit{throughput} over a set of tasks, capturing both how well and how fast each task is executed. Accordingly, each task is evaluated through two components: a \textit{correctness score} and a \textit{speed score}.

The correctness component reflects task completion quality and is normalized between 0 and 1. For the scaffolded manipulation tasks, correctness is defined as the fraction of progress achieved relative to the task goal. In the vertical configuration, this corresponds to the number of notches successfully traversed divided by the total number of notches. Similarly, in the horizontal configuration, correctness is the number of curves cleared divided by the total.
For the continuous rotation configuration, correctness is computed as the achieved rotation normalized by a full $360^\circ$ rotation.

Pure grasping tasks use a discrete scoring scheme: a score of 0 is assigned if the object is not lifted, 0.5 if the object is lifted but dropped during relocation, and 1 if the object is successfully grasped and relocated.

The speed component captures execution efficiency and is defined relative to a human baseline. Specifically, it is computed as the ratio between a baseline time and the time taken by the evaluated system to complete the task.
The baseline time is obtained as the average completion time of human participants, as measured in the user study described in \Cref{sec:human experiments}.

The overall task score is computed as a weighted combination of correctness and speed:
\begin{equation}
    \text{Score} = 0.8 \cdot \text{Correctness} + 0.2 \cdot \text{Speed}.
\end{equation}
This weighting reflects the current state of the field: as in-hand manipulation remains a challenging frontier in robotic hand development, reliable task completion is prioritized over execution speed. Consequently, a higher weight is assigned to correctness (0.8) to emphasize robustness, while still accounting for efficiency through a smaller contribution from speed (0.2).
Scores greater than 1 indicate superhuman performance, which may arise in future scenarios such as learning-based controllers achieving faster-than-human execution; therefore, the speed score is intentionally left unbounded.
\subsection{Real-World Benchmark Design}
\label{sec:real_world_design}

A central design goal of the POMDAR benchmark is reproducibility across laboratories. To this end, all physical components of the benchmark apparatus are manufactured using consumer-grade 3D printing, eliminating the need for specialized machining or procurement of proprietary parts. 

The system is secured to the table with clamps, while protrusions, bolt holes, and a custom bolt–nut retention mechanism ensure repeatable mounting and easy replacement of damaged parts without reprinting the full base. This makes the real-world setup compact, maintainable, and practical for repeated benchmarking.

\subsection{Simulation Benchmark Design}
\begin{figure}
    \centering
    \includegraphics[width=\linewidth]{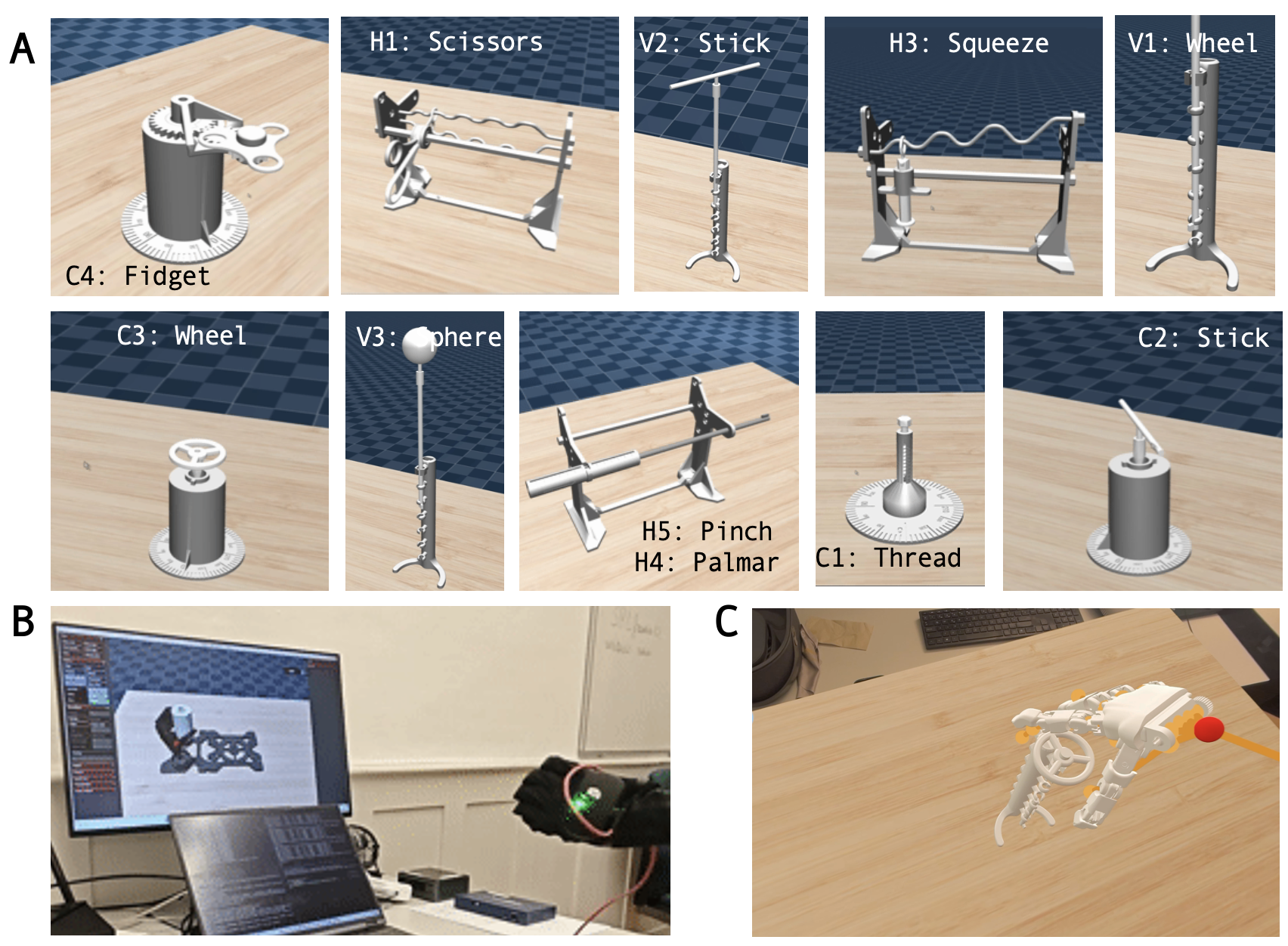}
    \caption{(A) Examples of the POMDAR tasks implemented in MuJoCo across different manipulation configurations. (B) Teleoperation using motion-capture gloves for direct control of the simulated hand. (C) Teleoperation in virtual reality using Apple Vision Pro, where the simulation is streamed to the user for an immersive and interactive experience; the tracked hand keypoints are visible as orange overlays.}
    \label{fig:sim_benchmark}
\end{figure}

The benchmark was implemented in MuJoCo with a task-oriented modeling strategy that preserves only the geometries essential for contact and scoring, while simplifying or discarding non-functional parts to keep the simulation efficient and stable. Because MuJoCo does not natively handle non-convex collision meshes well, complex components were either approximated with primitive shapes or preprocessed using CoACD for convex decomposition before import. Additional task-specific workarounds were required for mechanisms that are hard to reproduce directly, such as manually enforcing gear couplings and using a feedforward PID controller to emulate screw-thread motion.  
The simulated benchmark supports teleoperation through a variety of interfaces, including motion capture gloves and VR systems (see Fig.~\ref{fig:sim_benchmark}), which also provides an overview of all tasks implemented in MuJoCo. Beyond teleoperation, the simulation environment enables rapid evaluation of hand designs prior to fabrication and establishes a foundation for future learning-based approaches, where policies can be trained directly on the tasks, decoupling performance from operator skill and teleoperation setup. \label{sec:sim_design}

\section{Results}
\subsection{User Study}
\label{sec:human experiments}

We conducted a user study to validate the benchmark and establish human baseline performance. Six participants were asked to solve all tasks given only the start and goal configurations, without an explicit strategy. Five out of six participants were right-handed, and all performed the tasks with their right hands. Each participant completed three trials per task, yielding a total of 18 trajectories.

\begin{figure}
    \centering
    \includegraphics[width=0.95\linewidth]{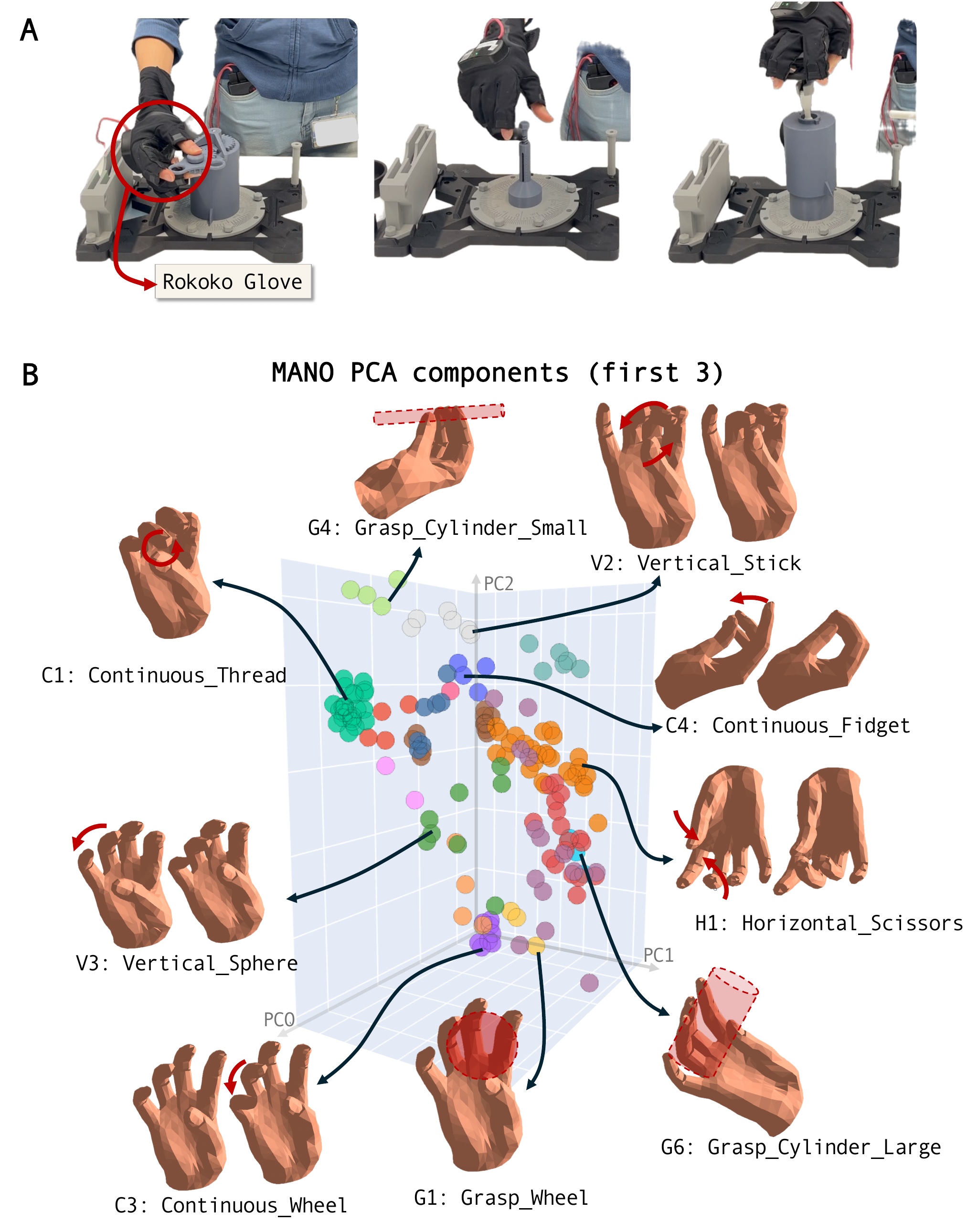}
    \caption{Human study and motion analysis.  (A) Example snapshots of the human data collection for three continuous rotation tasks. Participants perform the benchmark using a motion-capture Rokoko glove (circled), which tracks hand keypoints during task execution.  (B) Principal component analysis (PCA) of the recorded hand motions using the MANO representation. The first three principal components (out of six total) are shown. Each point corresponds to a 1.5\,s trajectory segment, and colors indicate different tasks. The clustering of points by color suggests that participants, both across and within subjects, adopt similar strategies for each task, indicating that the tasks are intuitive and well-constrained.  Example MANO reconstructions are shown around the plot, illustrating representative hand configurations. These are consistent with the intended manipulation patterns and grasp types derived from the taxonomy in Fig.~\ref{fig:manipulation_grasp_classification}.}
    \label{fig:human_study}
\end{figure}

Participants wore a motion-capture glove tracking 22 hand keypoints at 100\,Hz, enabling detailed analysis of hand motion (\Cref{fig:human_study}A). 

We verify that participants adopt motion and grasp patterns consistent with the taxonomies introduced in \Cref{sec: manipulation tasks consolidation} and \Cref{sec: grasping tasks consolidation}. The low variability in strategies suggests that the scaffolded design effectively constrains interactions to the intended motion primitives (\Cref{fig:human_study}B).

Baseline completion times are computed as the mean across all participants and trials, and they define the speed component of the benchmark score.

Finally, recorded hand trajectories are visualized in \Cref{fig:human_study}, showing consistent motion patterns across participants.
\subsection{Benchmark results}

We evaluate the POMDAR benchmark through teleoperation experiments using progressively more dexterous embodiments of the ORCA hand~\cite{christoph2025orcaopensourcereliablecosteffective}.

\begin{figure}
    \centering
    \includegraphics[width=1.0\linewidth]{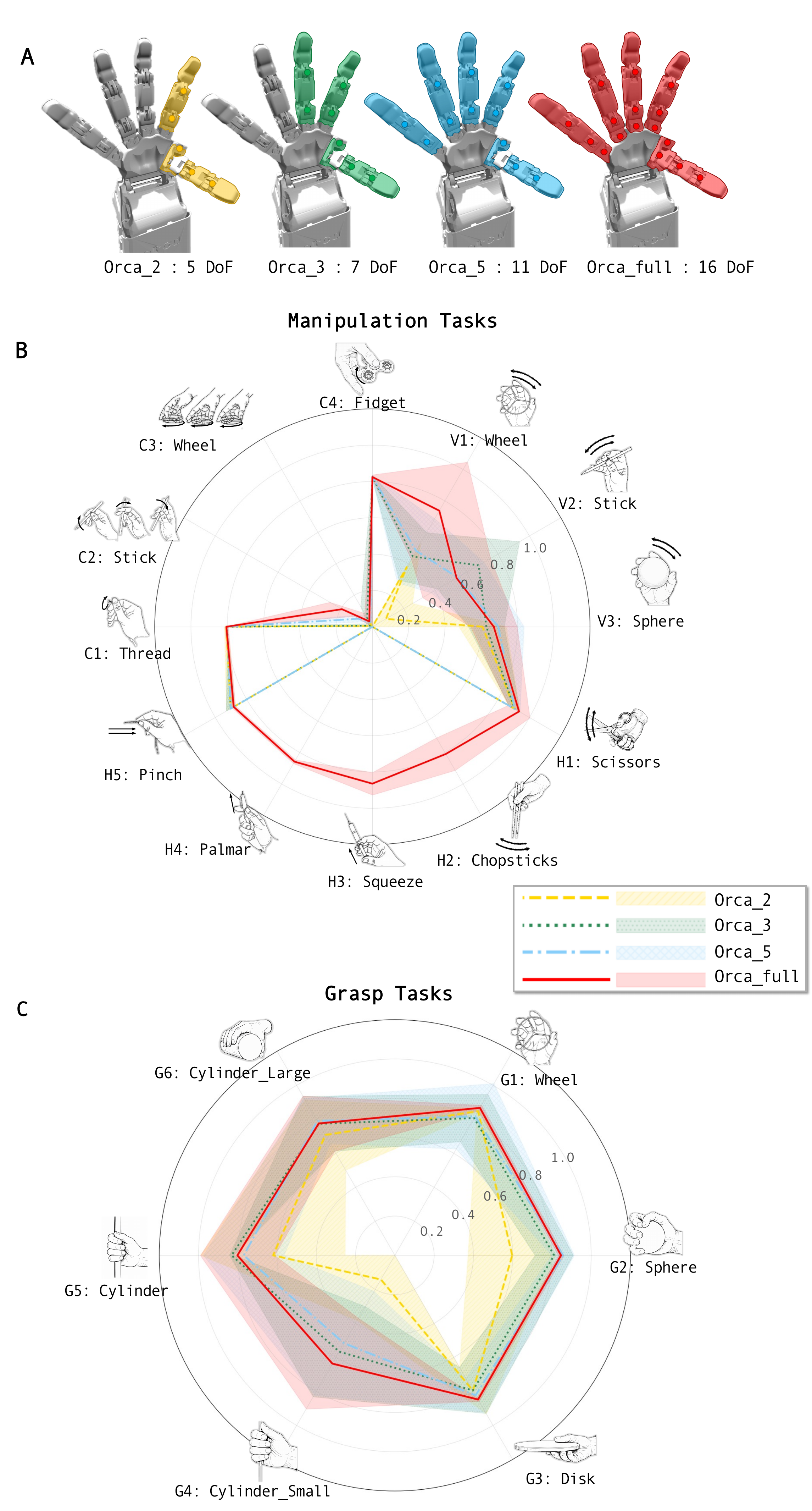}
    \caption{Results of the POMDAR benchmark across different ORCA hand embodiments. (A) The evaluated embodiments (Orca\_2, Orca\_3, Orca\_5, Orca\_full) are shown, where highlighted joints indicate the unlocked degrees of freedom. This selection of DoFs is specific to the ORCA hand and should not be interpreted as a general mapping between number of DoFs and achievable capabilities, but rather as a controlled study of embodiment variations within the same platform. All embodiments use the same controller and teleoperation interface; differences arise solely from locking selected DoFs at zero position in the hand controller. (B) Radar plots for the manipulation tasks (12 total), and (C) radar plots for the grasping tasks (6 total). Each task is repeated 20 times. The shaded regions represent the standard deviation across trials. Icons around the plots provide a visual reference for each task to aid interpretation. The color of each plot corresponds to the embodiment shown in (A); an explicit legend is omitted for clarity.}
    \label{fig:results_pomdar}
\end{figure}

Specifically, we consider four configurations: (i) a 2-finger setup (thumb and index, no abduction, 5 DoF), (ii) a 3-finger setup (no abduction), (iii) a 5-finger setup without abduction, and (iv) a full 5-finger configuration with 16 DoF (\Cref{fig:results_pomdar}A). The hand is mounted on a 7-DoF Franka Emika arm.
\paragraph{Robot Setup}
All experiments are conducted using the ORCA hand mounted with a $90^\circ$ adapter to a Franka Emika arm. The teleoperator wears Rokoko motion-capture gloves, which track 22 keypoints on the hand and forearm, as well as the 6-DoF wrist pose via a base station providing absolute tracking. The captured keypoints are mapped to the robot through a global scaling and rotation transformation to account for user-specific hand morphology. These parameters are automatically tuned via a Bayesian optimization procedure based on a small calibration dataset of seven poses (four pinch configurations, hand open with and without abduction, and closed hand), where glove keypoints are paired with ground-truth robot joint configurations.

The keypoints are then retargeted online to the robot joint space using an energy minimization-based algorithm (similar to~\cite{sivakumar2022robotictelekinesislearningrobotic}), running at approximately 25~Hz. The relative wrist pose is sent to the robot, which is controlled through a low-level impedance controller.

Depending on the task, the arm motion is either fixed (e.g., C4 fidget, H4 pinch, H5 palmar), constrained along the vertical axis (V1--V3, C1--C3), or constrained along the horizontal axis (H1--H3). The initial pose of the robot hand is selected by the operator and may vary across tasks and embodiments to compensate for alignment differences.
\paragraph{Experimental protocol.}

All experiments are performed by the same operator to ensure consistency. For each task and embodiment, the operator performs five practice trials to determine an effective strategy and a start pose for the robot, followed by 20 recorded trials.

\paragraph{Disclaimer.}
The reported results reflect the performance of a \emph{combined system} comprising the robotic hand, the teleoperation interface, and the operator's skill. While the results are directly comparable across the tested embodiments, they are not necessarily comparable to results obtained with different teleoperation systems. They nonetheless provide meaningful insights into the role of embodiment in dexterous manipulation. Moreover, the object dimensions are designed for anthropomorphic hands, which may disadvantage embodiments whose morphology deviates from human hand proportions, potentially leading to lower benchmark scores.  

\paragraph{Teleoperation Results.}
To test whether the benchmark provides useful insights into the dexterity of robotic hands and whether it correlates with intuitive measures such as the number of degrees of freedom, we evaluate four different embodiments (\Cref{fig:results_pomdar}A). These correspond to variants of the ORCA hand with selected degrees of freedom locked or unlocked. The configurations span from 5 to 16 DoF (full hand without wrist), where in the full configuration all finger abduction DoFs are also enabled.

\Cref{fig:results_pomdar}B reports results for vertical, horizontal, and continuous manipulation tasks. Performance differences are most pronounced in tasks requiring coordinated multi-finger interactions and finger abduction. 

Certain tasks are only feasible with specific embodiments, particularly those requiring thumb or multi-finger abduction (e.g., squeeze, palmar, and chopsticks). Conversely, several tasks exhibit identical execution strategies across embodiments (e.g., scissors, thread, and pinch), resulting in comparable performance and reduced sensitivity to embodiment differences. This highlights that dexterity gains are strongly task-dependent.

\Cref{fig:results_pomdar}C shows radar plots for the grasping tasks. All embodiments are able to successfully grasp the objects; however, increased dexterity improves both grasp stability and execution speed. This effect is most pronounced for smaller objects, where additional fingers significantly enhance robustness, while for larger objects the main benefit is faster relocation. Notably, increasing the number of fingers from three to five without abduction yields limited improvement, whereas adding a third finger (from two to three) provides a substantial performance gain, primarily due to improved grasp stability.

Task difficulty varies across configurations. Vertical tasks exhibit higher variance due to finer-grained correctness metrics and increased coordination requirements. In contrast, continuous rotation tasks are generally the most challenging, particularly those involving the gravity-based clutch, which cannot be reliably solved with the current hand and teleoperation system. Simpler tasks such as fidget and thread are easier, as they require fewer degrees of freedom and do not penalize temporary loss of contact.

In total, the benchmark results are based on 1140 recorded trajectories (18 tasks, 4 embodiments, 20 trials per task, excluding non-repeated cases), corresponding to approximately 25 hours of real-world testing. All experiments were conducted with the same robot, operator, and teleoperation interface, ensuring consistent comparison across embodiments.

Finally, \Cref{fig:teaser} (right) presents aggregated scores as a function of the number of DoF. While performance generally increases with dexterity, the results confirm that improvements are task-dependent, with some tasks benefiting from additional fingers and others from specific kinematic capabilities such as abduction. The error bars represent variability across trials: for each trial, we compute the average score over all tasks and report the mean and standard deviation across the two trials.

\begin{figure}[!t]
    \centering
    \includegraphics[width=1.05\linewidth]{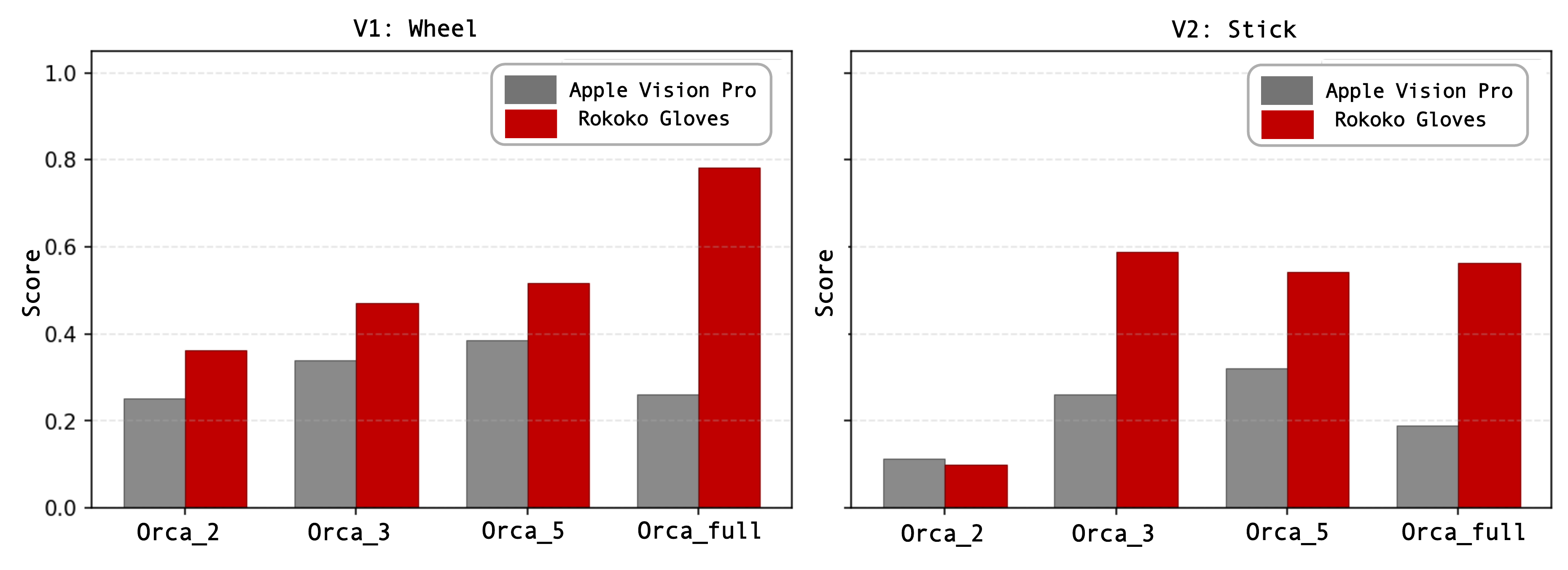}
    \caption{Comparison of teleoperation methods using motion-capture gloves (Rokoko) and Apple Vision Pro (AVP) across different ORCA embodiments for representative tasks (V1: Wheel, V2: Stick). Results are averaged over 20 trials per task. The teleoperation stack, controller, and robot setup are identical in both cases; the only difference lies in the source of hand keypoints. Performance with AVP is consistently lower, primarily due to occlusions in egocentric perception, which limit accurate tracking of finger motions during manipulation.}
    \label{fig:avp_rokoko}
\end{figure}

\paragraph{Ablations}
We demonstrate that the benchmark can also be used to compare different teleoperation methods and equipment. \Cref{fig:avp_rokoko} shows the results obtained using motion capture gloves and Apple Vision Pro (AVP) across all embodiments. Overall, the scores achieved with AVP are lower, mainly due to occlusions in egocentric perception. This is particularly evident in tasks such as the vertical configurations, where finger movements are partially occluded by the thumb and therefore not fully captured, leading to reduced control accuracy. 

\begin{figure*}[!t]
    \centering
    \includegraphics[width=0.90\textwidth]{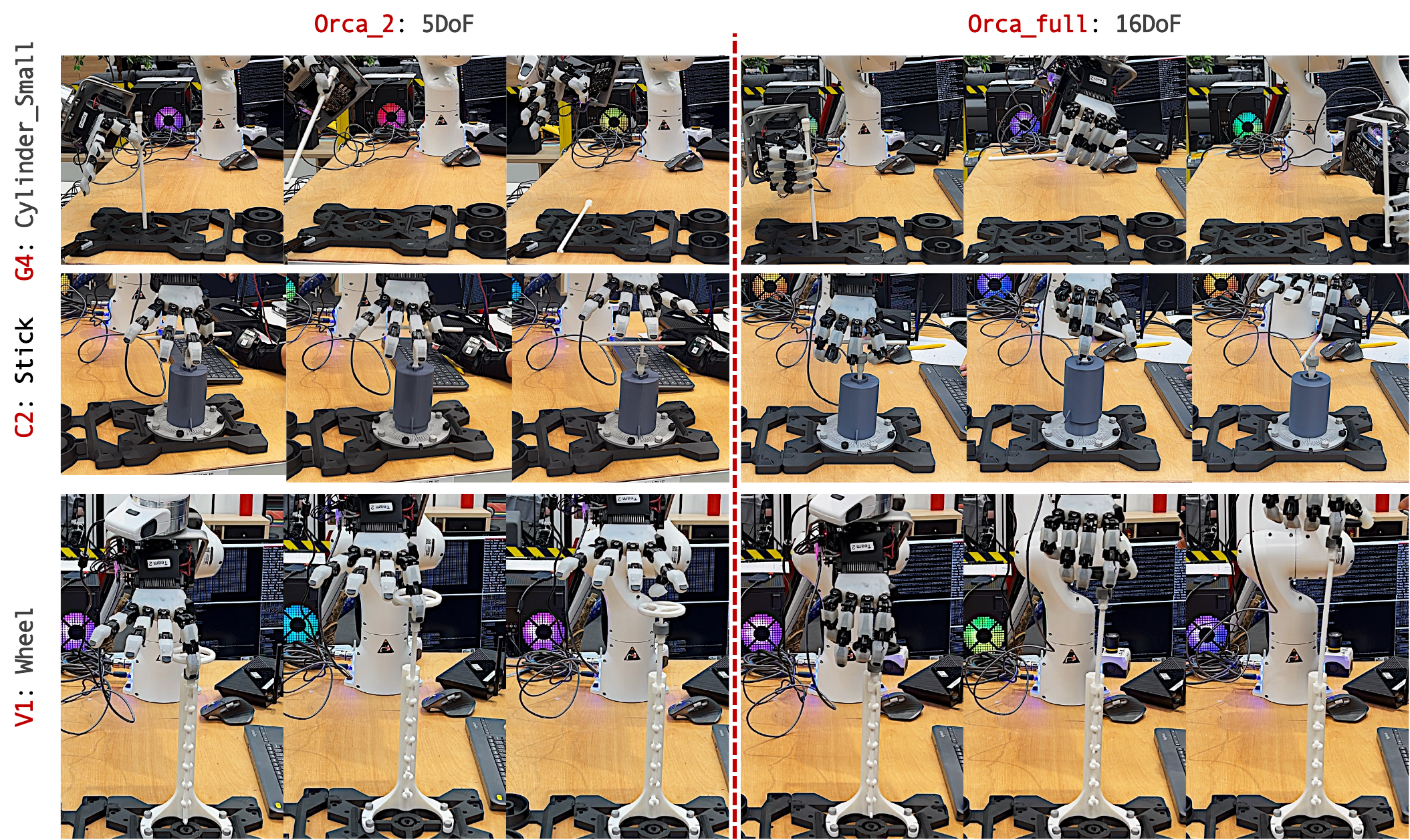}
    \caption{Qualitative example task execution sequences for the 5Dof (left) and 16 Dof (right) versions of the ORCA hand. The shown tasks where selected to highlight differences in the performance due to the added degrees of freedom. }
    \label{fig:sequences}
\end{figure*}
\section{Discussion}

The results demonstrate that POMDAR provides a quantitative, taxonomy-grounded framework for evaluating dexterous manipulation across robot hand embodiments. We outline key limitations and future directions.

\paragraph{Teleoperation-only evaluation.}
The current results reflect a combined system of the hand, teleoperation interface, and human operator, and therefore do not isolate mechanical dexterity. Evaluating autonomous policies, whether learned or programmed, would decouple hand capability from operator skill and enable more direct embodiment and control method comparisons.

\paragraph{Object size and anthropomorphic bias.}
The benchmark objects are designed for anthropomorphic hands, which may disadvantage embodiments with different morphologies. Future work could explore scalable or parameterized object sets to enable fairer comparisons.

\paragraph{Limited interaction dynamics.}
The benchmark primarily evaluates kinematic dexterity and controlled contact interactions, and does not capture dynamic behaviors relying on inertial or forceful interactions (e.g., pushing, tossing, flipping). Extending the task set in this direction would broaden the notion of dexterity.

\paragraph{Dependence on external robotic arm.}
The benchmark requires mounting the hand on an external robotic arm, which may introduce variability across systems. This is partially mitigated by constraining end-effector motion (fixed or single-axis) in most tasks. Future designs could explore standardized mounting or arm-independent setups.

\paragraph{Toward automated benchmarking.}
A key direction is the development of learning-based methods for automatic evaluation of robotic hands in simulation and reality. Recent work in dexterous retargeting, including sampling-based approaches~\cite{pan2026spiderscalablephysicsinformeddexterous} and reinforcement learning-based methods~\cite{mandi2025dexmachinafunctionalretargetingbimanual}, suggests that human motion can be transferred across embodiments. Such systems could generate task policies and compute benchmark scores without teleoperation. The simulation version of POMDAR provides a foundation for scalable and reproducible evaluation framework.
\section{Conclusion}
We presented POMDAR, a performance-based benchmark for evaluating dexterity in anthropomorphic robot hands. The benchmark is grounded in established manipulation and grasp taxonomies, translating 14 manipulation patterns and 33 grasp types into a structured set of 18 physical tasks covering both manipulation (12 tasks) and pure grasping (6 tasks).

POMDAR provides a quantitative dexterity score based on task correctness and execution speed relative to a human baseline, enabling direct and interpretable comparisons across embodiments.

A key design principle is reproducibility: all components are fully 3D-printable and open-source, and a simulation counterpart enables evaluation without physical hardware.

We validated the benchmark through a user study and experiments across multiple ORCA hand embodiments (5–16 DoF), showing that POMDAR scores track embodiment dexterity and reveal task-dependent performance differences.



\section*{ACKNOWLEDGMENT}
D.L is supported by Swiss National Science
Foundation (SNSF) Project Grant No. 200021 215489. Y.T. received support from the Takenaka Scholarship Foundation, Max Planck ETH Center for Learning Systems, and the Swiss Government Excellence Scholarship.
The authors thank Harry Durham for assistance with the hardware design of the benchmark. 
\FloatBarrier
\balance
\bibliography{references, web_references, hands_web_references}
\bibliographystyle{IEEEtran}
\end{document}